\documentclass[letterpaper]{article} 
\usepackage[]{aaai24}  
\usepackage{times}  
\usepackage{helvet}  
\usepackage{courier}  
\usepackage[hyphens]{url}  
\usepackage{graphicx} 
\usepackage{natbib}  
\usepackage{caption} 
\usepackage{amssymb}
\usepackage{pifont}
\usepackage{algorithm}
\usepackage{algorithmic}
\usepackage{amssymb}
\usepackage{amsmath}
\usepackage{mathtools}
\usepackage{amsfonts} 
\usepackage{fixmath}

\usepackage{graphicx}
\usepackage[utf8]{inputenc} 
\usepackage[T1]{fontenc}    
\usepackage{url}            
\usepackage{booktabs}       
\usepackage{amsfonts}       
\usepackage{nicefrac}       
\usepackage{microtype}      
\usepackage{xcolor}         
\usepackage{blindtext}
\usepackage{multirow}

\usepackage{listings}
\DeclareCaptionStyle{ruled}{labelfont=normalfont,labelsep=colon,strut=off} 
\floatstyle{ruled}
\newfloat{listing}{tb}{lst}{}
\floatname{listing}{Listing}
\usepackage{multirow}
\usepackage{amsmath}
\usepackage{xcolor}
\usepackage{lipsum}

\title{SupplyGraph: A Benchmark Dataset for Supply Chain Planning using Graph Neural Networks}

\author{
    Azmine Toushik Wasi,
    MD Shafikul Islam,
    Adipto Raihan Akib
}
\affiliations{
    Shahjalal University of Science and Technology, Bangladesh \\
    azminetoushik.wasi@gmail.com, shafikul37@student.sust.edu, aqibadipto27@gmail.com
    
}

\begin{document}

\maketitle


\begin{abstract}
Graph Neural Networks (GNNs) have gained traction across different domains such as transportation, bio-informatics, language processing, and computer vision. However, there is a noticeable absence of research on applying GNNs to supply chain networks. Supply chain networks are inherently graph-like in structure, making them prime candidates for applying GNN methodologies. This opens up a world of possibilities for optimizing, predicting, and solving even the most complex supply chain problems. A major setback in this approach lies in the absence of real-world benchmark datasets to facilitate the research and resolution of supply chain problems using GNNs. To address the issue, we present a real-world benchmark dataset for temporal tasks, obtained from one of the leading FMCG companies in Bangladesh, focusing on supply chain planning for production purposes. The dataset includes temporal data as node features to enable sales predictions, production planning, and the identification of factory issues. By utilizing this dataset, researchers can employ GNNs to address numerous supply chain problems, thereby advancing the field of supply chain analytics and planning.
Source: \textcolor{blue}{\texttt{https://github.com/CIOL-SUST/SupplyGraph}}

\end{abstract}

\section{Introduction}
\textbf{Supply chain} is a dynamic network of organizations that participate in the various processes and activities that produce value in the form of products and services for consumers via upstream and downstream linkages \cite{mentzer2001defining}, entailing a continuous flow of information, goods, and money among its various stages \cite{higuchi2004dynamic}. 
As supply chain consists of various entities interconnected in a complex network, it involves intricate inter-dependencies and complex decision-making processes. Besides, the modern supply chain produces an enormous amount of data; relationships and dependencies between all entities are not straightforward and require sophisticated models to capture \cite{sadeghiamirshahidi2014improving, chaovalitwongse2013computational}. 

\begin{figure*}[t] 
\centering {\includegraphics[scale=0.74]{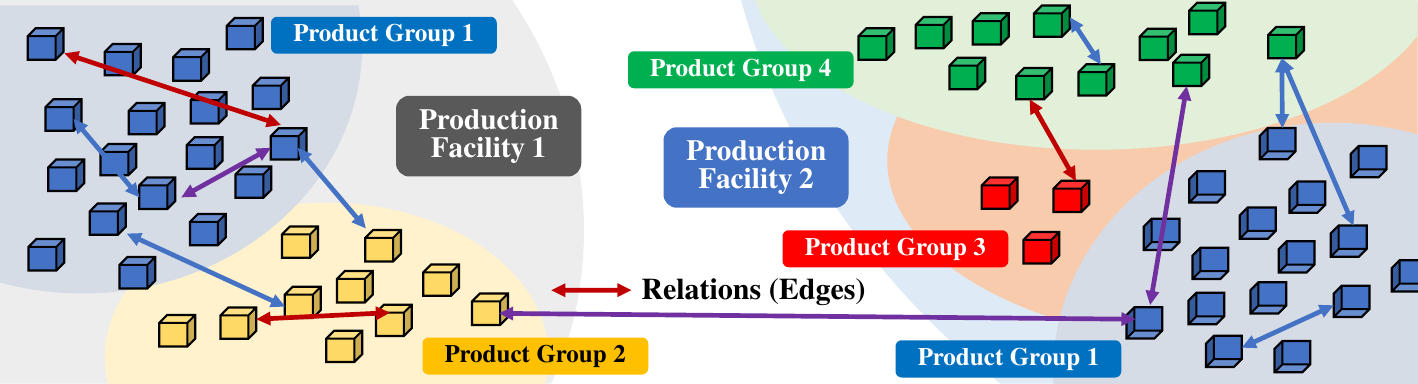}}
\caption{\textbf{Supply Chain Problem Formulation in Homogeneous Graph.} Boxes represent various product types, with color indicating different groups. They are closely located based on product groups and production facilities. Different relational connections denote shared raw material requirements, interdependence between products, and other impacts.}\label{fig:1-formulation}
\end{figure*}

Production planning is a crucial component of supply chain management as predicting the future demand for products or services, which helps organizations optimize their inventory levels, production schedules, and resource allocation \cite{aamer2020data, zougagh2020prediction}. 
Company revenue greatly relies on accurately predicting demand and planning accordingly which has led to the exploration of various deep learning and machine learning models \cite{alves2021applying} \cite{pirhooshyaran2020simultaneous} \cite{pacella2021evaluation} to tackle this issue. However, GNNs can model network-like structures, such as trade flows in global trade or social networks, allowing for a better understanding of complex supply chain dynamics \cite{kosasih2022machine}. {An illustration explaining Supply chain as a graph is presented in Figure \ref{fig:1-formulation}.}  Although few studies have been conducted on applying GNNs to supply chain contexts only to predict hidden link prediction to mitigate risk \cite{kosasih2022machine} and uncover hidden dependency links in supply chains \cite{aziz2021data}. The absence of publicly available datasets in this domain obstructs progress and benchmarking. By furnishing a dedicated dataset, we bridge the gap between supply chain dynamics and the potential of GNNs, enabling novel models that can facilitate robust supply chain strategies for businesses.

\section{Related Works}
\subsection{Supply Chain and Machine Learning}
In the domain of production planning within supply chain management, extensive research has been dedicated to harness machine learning \cite{feizabadi2022machine, zhu2021demand, aamer2020data}. Numerous studies aim to improve demand prediction and optimize production processes through these methodologies \cite{nitsche2021mapping, younis2022applications, filali2021exploring}. For instance, deep learning techniques including Artificial Neural Networks (ANNs)  \cite{lunardi2021comparison,mrad2019improved,vairagade2019demand},  Convolutional Neural Networks (CNNs)  \cite{husna2021demand,tang2022cnn,li2021algorithm}, and Long Short-Term Memory networks (LSTMs) \cite{bousqaoui2021comparative, lingelbach2021demand}.  Researchers have increasingly leveraged these methods to enhance the efficiency and effectiveness of production planning and supply chain management.

\begin{table*}[h]
\centering
\caption{Dataset Information} \label{tab:datasets}
\begin{tabular}{lrlrlr}
\toprule
\multicolumn{2}{c}{ Node }  & \multicolumn{2}{c}{ Edge Classes }  & \multicolumn{2}{c}{ Edge Count } \\
\hline
Total Products (Nodes)& 41      & Total Edge Types   & 62   & Total Unique Edges  & 684 \\
Number of Groups      & 5       & Class (Group)      & 5    & Count (Group)      & 188\\
Number of Sub-groups  & 19      & Class (Sub-group)  & 19   & Count (Sub-group)  & 52\\
Number of Plants      & 25      & Class (Plant)      & 25   & Count (Plant)      & 1647\\
Number of Storage    & 13      & Class (Storage)    & 13   & Count (Storage)    & 3046\\
\bottomrule
\end{tabular}
\end{table*}

\begin{figure*}[h] 
\centering {\includegraphics[scale=0.5]{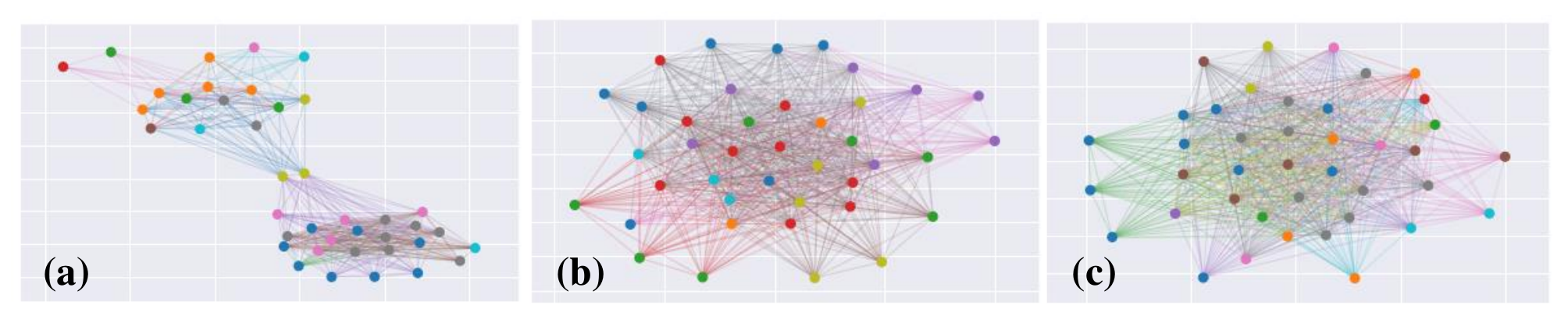}}
\caption{Homogeneous Graphs from the dataset. (a) Nodes are sub-group products, and plants are edges. (b) Nodes plant-products and storage locations are edges. (c) Nodes are sub-group products and storage locations are edges.}\label{fig:3Graphs}
\end{figure*}

\subsection{Supply Chain and Graph Neural Networks}
Supply Chain machine learning with Graph Neural Networks holds significant promise by enabling the modeling of complex supply chain structures, optimizing logistics, and enhancing decision-making through data-driven insights. Recent advancements in this field leverage GNNs to improve demand forecasting, and enhance supply chain resilience, paving the way for more efficient and adaptive supply chain operations.
Also, the utilization of graph representation learning methodologies has led to breakthroughs in revealing hidden dependency links within supply chain networks, further elevating the performance of link prediction tasks \cite{aziz2021data, kosasih2022machine} inspired by previous GNN link prediction works \cite{zhang2018link, Zhang2017WeisfeilerLehmanNM, Cai2020LineGN, Cai2020AMA}. This research highlights machine learning's effectiveness in refining demand forecasting and production planning, with GNNs showing potential in addressing supply chain challenges. 
However, growth in GNN-based supply chains research is very slow. It shows the significance of specialized datasets and cutting-edge methodologies for fully utilizing GNNs' advantages in addressing the complex demand prediction and production planning issues that arise in the supply chain domain.

\section{Data Collection}
We obtained data from the central database system of one of the biggest and most prominent FMCG (Fast Moving Consumer Goods) companies in Bangladesh. We reorganize it for temporal graph utilization, extracting nodes and features. We are unable to publish the actual product names, product codes, and the name of the company, as doing so could potentially compromise the competitive standing of the concerned company. Table \ref{tab:datasets} summarizes the dataset information.  \\
\textbf{Quality Control.} Each individual node, edge, and node feature undergoes a thorough manual examination and validation process. This involves scrutinizing the data for any irregularities, such as anomalies or missing information. Additionally, even zero values are meticulously reviewed to ensure their accuracy and legitimacy.  \\
\textbf{Temporal \& Location Coverage.} The occurrences in this database cover the period from January 1, 2023, to August 9, 2023, and include four features: production, sales orders, delivery to distributors, and factory issues in two metrics: unit and weight produced in metric tons. The company operates across the entirety of Bangladesh.

\subsection{Feature Description}
Within SupplyGraph, \textbf{nodes} pertains to distinct products, while \textbf{edges} represent various connections linking these products: same product group or sub-group, same plant or storage location. Figure \ref{fig:3Graphs} presents some examples of homogeneous graphs.\\
In the temporal data, node features include production, sales orders, delivery to distributors, and factory issues. \\
\textit{Production}, which quantifies product output considering sales orders, customer demand, vehicle fill rate, and delivery urgency. This quantity is typically measured in units or Metric Tons. 
\textit{Sales Order} signifies distributor-requested quantities, pending approval from the accounts department. It reflects overall product demand. 
\textit{Delivery to Distributors} denotes dispatched products aligning with orders, impacting company revenue significantly. 
\textit{Factory Issue} covers total products shipped from manufacturing facilities, with some going to distributors and the rest to storage warehouses. 
We have all the temporal data in two modes: number of units produced (example: 500 units) and total weight example: 10 metric tons) produced.

\subsection{Data Statistics}
\subsubsection{Product Group and Sub-groups}

\begin{figure}[h] 
\centering {\includegraphics[scale=0.2]{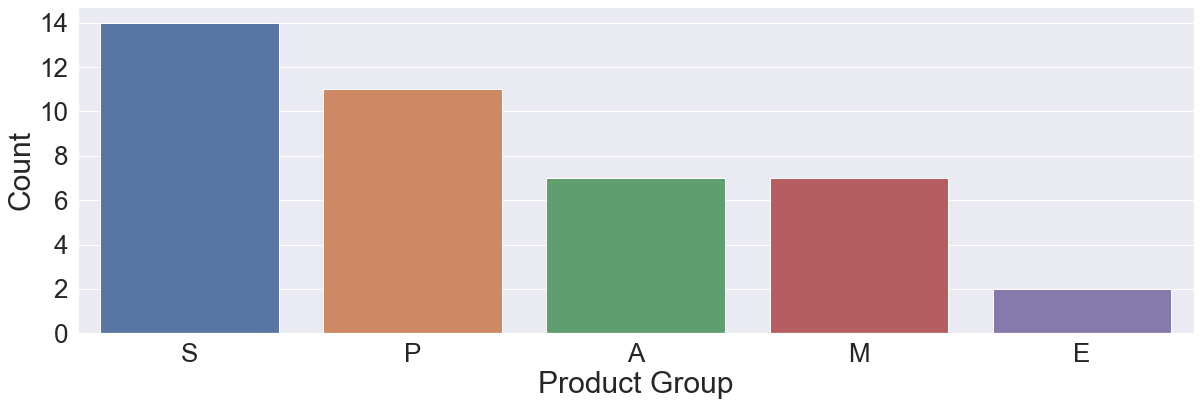}}
\caption{Product group count.}\label{fig:PG-Count}
\end{figure}

\begin{figure}[h] 
\centering {\includegraphics[scale=0.2]{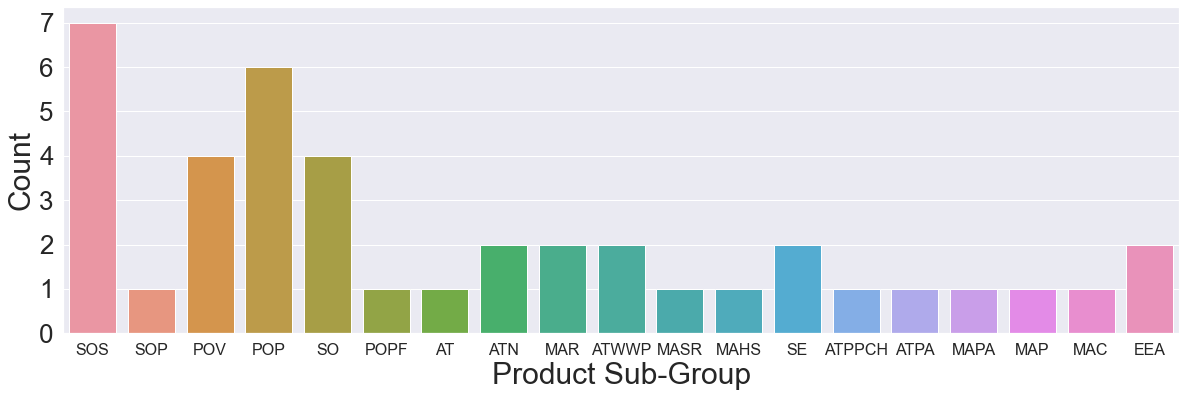}}
\caption{Product subgroup count.}\label{fig:SPG-Count}
\end{figure}

Figure \ref{fig:PG-Count} shows that product group “S” has the highest number of products reflecting its highest variety and product group “E” has the lowest number of products. Notably, a majority of the product groups exhibit a range of around 8 to 10 products each.
Figure \ref{fig:SPG-Count} reveals the diversity within product sub-groups. As previously noted, the "s" product group stands out for its remarkable variety, and this characteristic is consistently maintained within its corresponding sub-group. Among our product subgroups, "SOS" emerges as the one encompassing the largest array of product categories. Interestingly, despite the elevated diversity exhibited by the product group, the corresponding sub-group maintains a relatively stable count.

\subsubsection{Analyzing Temporal Trends.}
\begin{figure}[h] 
\centering {\includegraphics[scale=0.33]{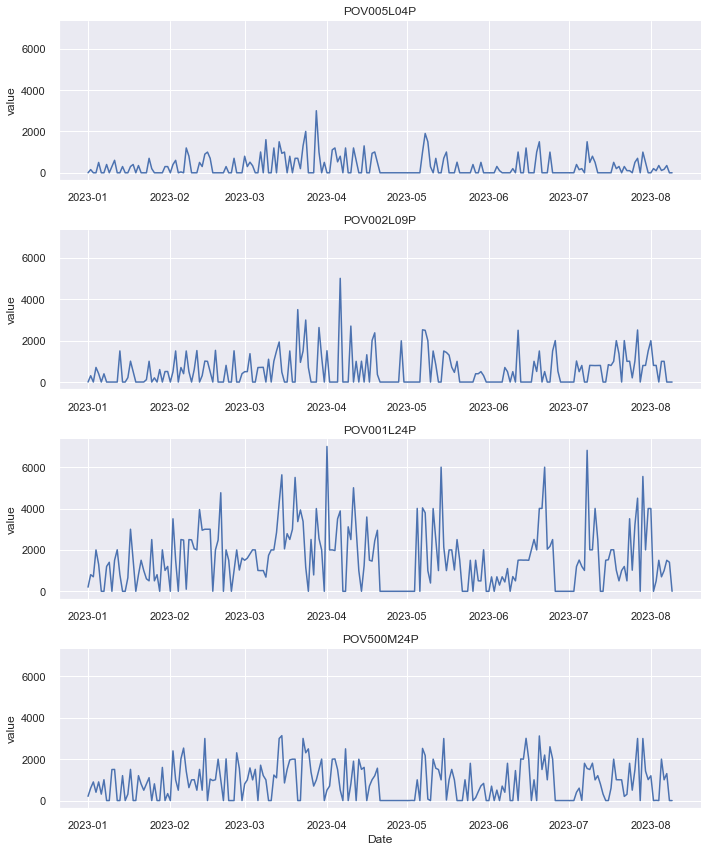}}
\caption{All product's production (MT) of POV sub-group.}\label{fig:POV Sub-group}
\end{figure}

Figure \ref{fig:POV Sub-group}, A particularly attractive pattern emerges: the production curve exhibits distinct and sharp fluctuations. Notably, the production unit of the company occasionally closes, which can be attributed to factors such as scheduled vacations, adjustments in the product's MRP rate, demand variations, and optimizations in transportation policies. In contrast, the other aspects of the graph, encompassing sales orders, delivery to distributors, and factory issuance, remain notably stable and consistent. Their minimal deviations stand in contrast to the variabilities characterizing the production curve, providing an insight into the dynamics of the supply chain activities.

\begin{figure}[h] 
\centering {\includegraphics[scale=0.33]{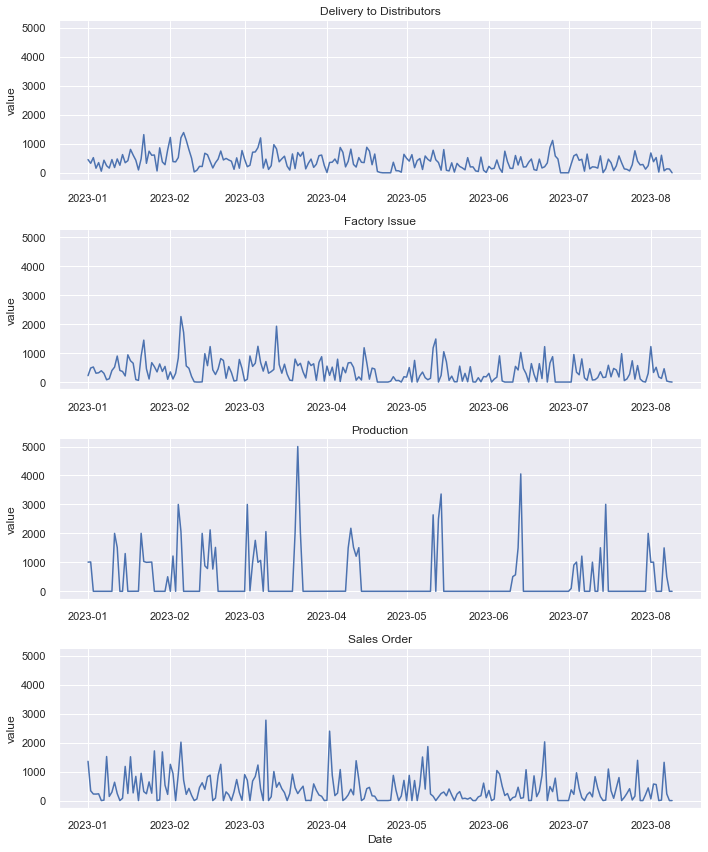}}
\caption{All 04 temporal values of product SOS008L02P.}\label{fig:SOS008L02P - SKU}
\end{figure}

In, Figure \ref{fig:SOS008L02P - SKU}, we can notice a very interesting pattern. The product is manufactured in large batches, with orders being accumulated over several days before production begins. The processes of delivering to distributors, handling factory issues, and fulfilling sales orders appear to be running smoothly and consistently. The company has opted to operate the production plant intermittently and maintain product inventory. When you observe straight lines in the production graph, it indicates that the plant is not actively engaged in production during those periods.

\subsubsection{Analyzing Temporal Correlations.}
\begin{figure}[h] 
\centering {\includegraphics[scale=0.52]{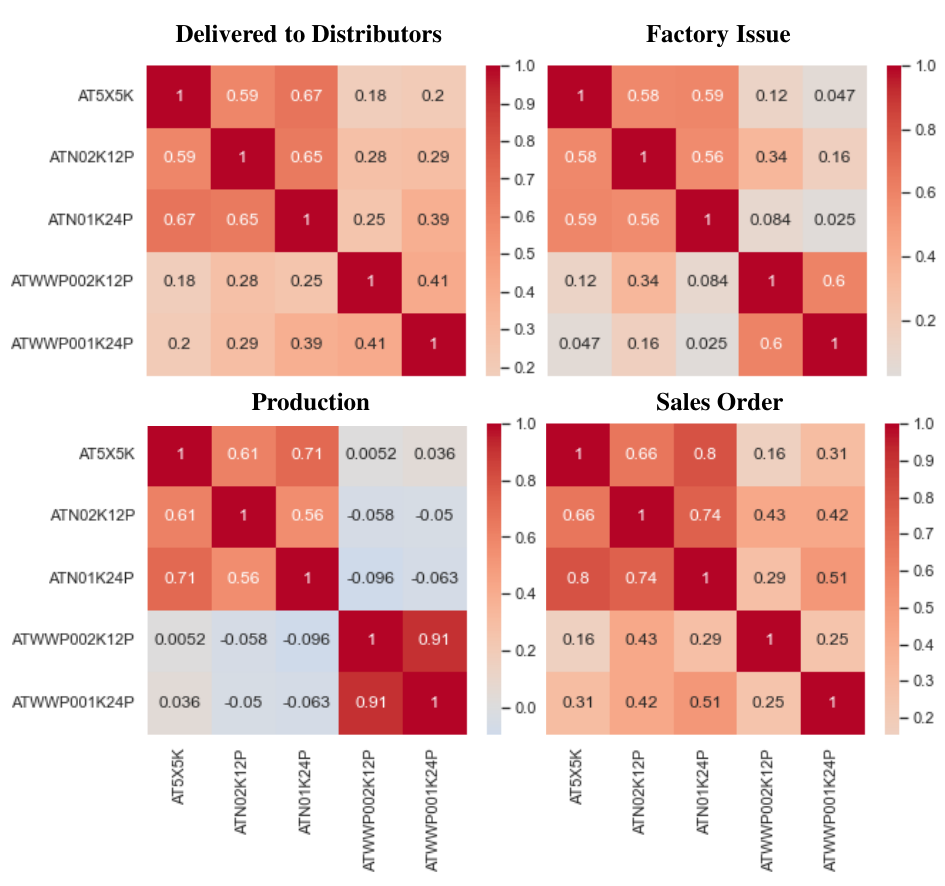}}
\caption{Temporal correlation of some products from Product Group A.}\label{fig:G-A-Corr}
\end{figure}

The correlation plot portrayed in Figure \ref{fig:G-A-Corr} serves to demonstrate the ways in which diverse subgroups of product "A" are interconnected in a multitude of facets. It is worth noting that the relationship between delivery to distributors and sales orders is particularly robust, whereas the production unit exhibits a distinct and prominent correlation pattern. Furthermore, a noteworthy observation pertains to the correlation between the production quantities of two subgroups, specifically "ATWWPOO2K12P" and "ATWWPOO1K24P". Additionally, it is worth highlighting that factory issues and sales orders similarly display correlation trends, albeit with some discernible variations.

\begin{figure}[h] 
\centering {\includegraphics[scale=0.52]{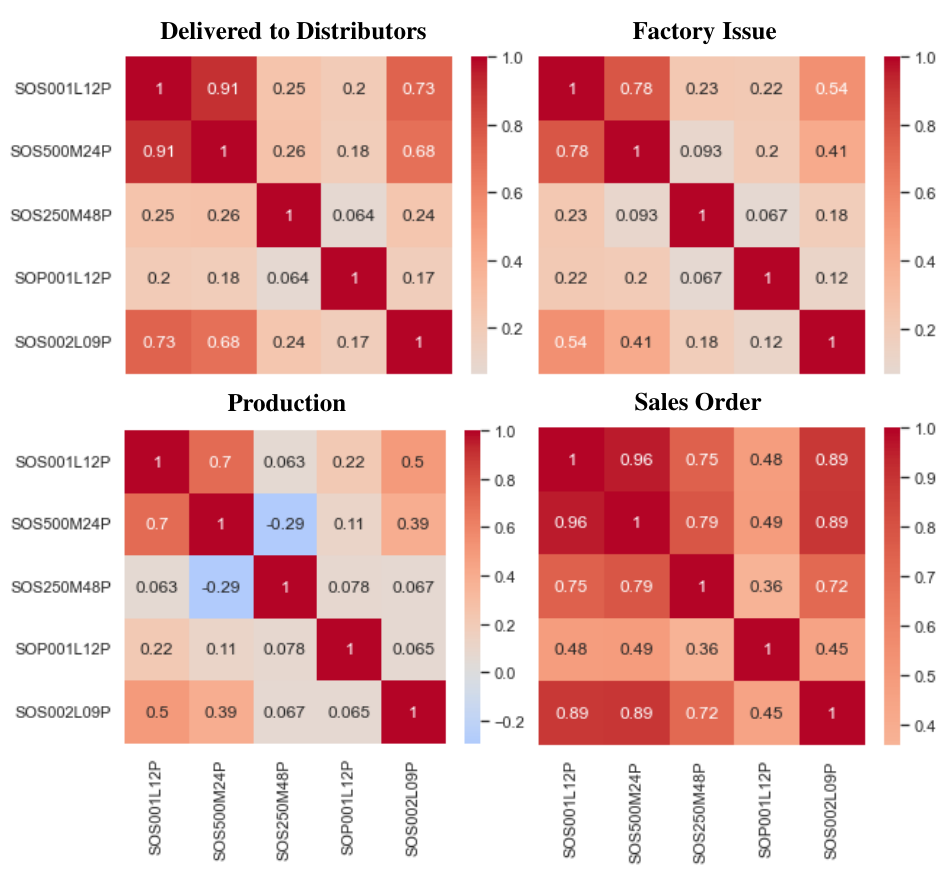}}
\caption{Temporal correlation of some products from Product Group A.}\label{fig:G-S-Corr}
\end{figure}

Upon examining the correlation plot of the "S" subgroup in Figure \ref{fig:G-S-Corr}, a distinct pattern becomes apparent: there is a significant and consistent correlation between the "SOSOO1L12P" and "SOS500M24P" subgroups across all variables. Notably, an intriguing observation arises in the context of sales orders, where a majority of the products within the "S" subgroup demonstrate a relatively higher correlation compared to delivery to distributors, factory issuances, and production.

The comprehensive analysis of correlations and temporal activities strongly indicates a substantial opportunity for the application of graph-based machine learning methodologies in the realm of supply chains. Leveraging the inherent structures and relationships within supply chain networks, graph-based machine learning holds great promise in enhancing decision-making processes, optimizing resource allocation, and fostering adaptability in the dynamic landscape of supply chain operations. 

\section{Dataset Potential and Practical Applications}
Our dataset's structural design models products as nodes, while the interrelationships, such as shared product groups, subgroups, production facilities, and storage locations, are represented as edge connections. This dataset equips us to apply GNNs to complex supply chain predicaments, such as: \\
\textbf{Supply chain planning} involves the utilization of historical data pertaining to production and demand as node attributes, with the integration of time-related features to capture seasonality and trends. To enhance demand forecasting accuracy, related entities' influence can be harnessed through neighborhood nodes. The hierarchical structure of nodes, including product groups, subgroups, plants, and storage locations, can be used for hierarchy-aware forecasting. In data-scarce scenarios, improved forecasting can be achieved by applying transfer learning across nodes of the same classification.\\
\textbf{Product Classification} involves classifying products by product groups, sub-groups, facilities; and also by economic profitability and production similarities for economic decision-making.\\
\textbf{Product Relation Classification} involves classifying or predicting product relations in the supply chain graph. Product Relation detection involves detecting a missing edge or relation in the supply chain graph by a binary prediction objective.\\
\textbf{Supply chain optimization} involves modeling goods flow between nodes, considering demand, lead times, and reorder points. Optimal routes and quantities can be recommended using edges as transportation links. Nodes representing plants can also aid in suggesting production adjustments based on demand projections, constraints, and capacity. \\
\textbf{Anomaly detection} is achieved by contrasting predicted demand with actual data to identify disruptions, stockouts, or unusual demand spikes. Deviations in demand from aggregated neighborhood demand can signal inconsistencies within neighborhoods, which can be detected as anomalies.\\
\textbf{Event classification} involves the training of GNNs to classify events based on edges and nodes, such as production capacity changes, new product launches, and disruptions. \\
\textbf{Fluctuation Detection} entails training GNNs to identify and classify fluctuations in the network based on impacts of global price hikes, supply chain issues, and disruptions. Fluctuations in supply chains refer to disruptions that can occur due to various factors and can significantly impact the operations of businesses. These disruptions can amplify negative shocks, affecting not just the firm experiencing the failure, but also its suppliers and customers, and even firms in other parts of the production network \cite{NBERw27565}. Using temporal data, the model can recognize patterns in these fluctuations over time and determine if any demand and supply fluctuation can occur in near future for better planning. \\
\textbf{Combinatorial} \cite{cappart2022combinatorial} and \textbf{Constrained} \cite{kotary2021endtoend} \textbf{Optimization} involves utilizing GNNs to optimize complex decision-making within the supply chain, considering various factors and constraints. GNNs can effectively model the intricate relationships between nodes and edges, enabling efficient allocation of resources, route planning, and inventory management, ultimately leading to improved  performance.


\textbf{Heterogeneous Graphs.} 
All these tasks can be executed in both heterogeneous and homogeneous forms. In the context of the heterogeneous approach, the incorporation of nodes representing distinct entities, such as products, storage locations, and production facilities, interconnected by edges that signify their intricate relationships, becomes a pivotal strategy. This approach facilitates the construction of a heterogeneous graph model, which effectively encapsulates the intricate web of interactions inherent in supply chain networks.
Embracing these heterogeneous models affords us the capacity to encompass the varied characteristics of different supply chain components and their complex interdependencies. This approach proves instrumental in addressing the multifaceted challenges that characterize supply chain management. The ability to comprehensively represent and analyze these heterogeneous relationships equips decision-makers with a deeper understanding of the dynamics within the supply chain, empowering them to devise optimized strategies and adapt to the dynamic nature of supply chain operations. In doing so, overall supply chain performance and resilience can be significantly enhanced.

\textbf{Hypergraphs.} 
Efficient task execution can also be achieved through hypergraph models. Nodes, representing entities like products, storage locations, and production facilities, intricately interconnect through hyperedges, capturing the intricate relationships within supply chain networks. This strategic use of hyperedges facilitates the construction of a tailored hypergraph model, adept at encapsulating the complex web of interactions inherent in supply chain networks.
The adoption of hypergraph models can provide a dynamic framework to address the diverse characteristics and intricate interdependencies of supply chain components, serving as a pivotal strategy for managing the multifaceted challenges inherent in supply chain management and fostering a deeper understanding of its dynamics, which can contribute to the enhancement of overall supply chain performance and resilience.

\section{Conclusion}
Supply chain networks present a potent framework for optimization and problem-solving for GNNs. However, the lack of real-world benchmark datasets impedes progress in this field. To address this issue, our work aims to provide a real-world benchmark graph-based dataset from a prominent FMCG firm in Bangladesh, with a focus on production-oriented supply chain planning. This dataset includes interconnected product nodes, as well as temporal features for sales prediction, production planning, and issue identification. Experimental results establish the importance of GNN-based methods in supply chains. Researchers can utilize this dataset to leverage graph neural networks in resolving multifaceted supply chain issues, thus catalyzing advancements in supply chain analytics and planning.

\section{Acknowledgement}
We thank the reviewers for their thoughtful comments and feedback. We appreciate all the support from the Computational Intelligence and Operations Lab, SUST. 

\clearpage
\newpage
\bibliography{aaai24}

\end{document}